\documentclass{article}

\PassOptionsToPackage{numbers, sort, compress}{natbib}

\usepackage[preprint]{neurips_2023}

\usepackage[utf8]{inputenc} 
\usepackage[T1]{fontenc}    
\usepackage{hyperref}       
\usepackage{url}            
\usepackage{booktabs}       
\usepackage{amsfonts}       
\usepackage{nicefrac}       
\usepackage{microtype}      
\usepackage{xcolor}         

\usepackage{graphicx}
\usepackage{amsmath}
\usepackage{enumitem}

\title{Decoupled Local Aggregation for \\ Point Cloud Learning}

\author{%
  Binjie Chen \\
  Xiamen University \\
  \texttt{chenbinjie@stu.xmu.edu.cn} \\
  \And
  Yunzhou Xia \\
  Xiamen University \\
  \texttt{23020221154127@stu.xmu.edu.cn} \\
  \And
  Yu Zang\thanks{Corresponding Author} \\
  Xiamen University \\
  \texttt{zangyu7@126.com} \\
  \And
  Cheng Wang \\
  Xiamen University \\
  \texttt{cwang@xmu.edu.cn} \\
  \And
  Jonathan Li \\
  University of Waterloo \\
  \texttt{junli@uwaterloo.ca} \\
}

\begin{document}

\maketitle

\begin{abstract}
The unstructured nature of point clouds demands that local aggregation be adaptive to different local structures.
Previous methods meet this by explicitly embedding spatial relations into each aggregation process.
Although this coupled approach has been shown effective in generating clear semantics, aggregation can be greatly slowed down due to repeated relation learning and redundant computation to mix directional and point features.
In this work, we propose to decouple the explicit modelling of spatial relations from local aggregation.
We theoretically prove that basic neighbor pooling operations can too function without loss of clarity  in feature fusion,
so long as essential spatial information has been encoded in point features.
As an instantiation of decoupled local aggregation, we present DeLA, a lightweight point network, where in each learning stage
relative spatial encodings are first formed, and only pointwise convolutions plus edge max-pooling are used for local aggregation then.
Further, a regularization term is employed to reduce potential ambiguity through the prediction of relative coordinates.
Conceptually simple though, experimental results on five classic benchmarks demonstrate that DeLA achieves state-of-the-art performance with reduced or comparable latency.
Specifically, DeLA achieves over 90\% overall accuracy on ScanObjectNN and 74\% mIoU on S3DIS Area 5.
Our code is available at \url{https://github.com/Matrix-ASC/DeLA}.
\end{abstract}

\section{Introduction}

3D point clouds are suitable for tasks like scene understanding and modeling, as they naturally contain detailed spatial cues. 
Compared to images, graphs or other data representations, one distinct characteristic of point clouds is that they are unorganized.
Therefore, if a model is to extract fine patterns from a point cloud without organizing it, 
special care must be taken to follow its unordered and unstructured nature.
One core component of point-based models that needs to be tailored is local aggregation.

Local aggregation is a general term for operations that integrate information from neighboring points to learn the local context.
An essential precondition for local aggregation is the awareness of exact spatial relations, since otherwise points sharing the same neighbors might end up receiving the same signal, leading to confusion about their identities.
For that, traditional methods\cite{PointNet++, PointCNN, RSCNN, KPConv, PosPool, PointTransformer} model relative coordinates explicitly along the integration process.
By emphasizing spatial relations in each aggregation, ambiguity is well reduced.
However, this coupled approach requires mapping raw coordinates to internal representations and 
fusing this spatial information into point features in each aggregation. 
Either could bring forth nonnegligible complexity.
If relative coordinates are incorporated in an over-simplified way, then model's expressivity may very likely be compromised.

To fundamentally contain this complexity, we propose to decouple explicit relation learning from local aggregation.
A key step in achieving this is to introduce a non-directional substitute, which we call spatial encoding.
This encoding is designed capable of retrieving relative coordinates at a local level.
Though absolute coordinates are a valid option,
we discuss and opt for a relative alternative that's translation invariant while containing more local details. 
If spatial encoding has been embedded in point features, 
complex aggregation functionalities can theoretically be carried out by basic pooling and pointwise operations,
given that  these operations adhere to a specific pattern:
feature transformations and pooling operations are interleaved,
and the source feature after aggregation is preserved.

The analysis on decoupled local aggregation leads us to the design of DeLA, which is comprised of several learning stages. 
At each stage, after subsampling, a small PointNet processes relative coordinates of neighbors at that scale,
whose output serves as the spatial encoding.
Then interleaved pointwise MLPs and edge max-pooling are stacked for actual local aggregation.
To encourage better utilization of the spatial information, 
the network is regularized to predict relative coordinates with points' neighbors at the end of each stage.
It should be noted that the relative coordinates we encode form an unordered set,
while the ones we try to extract must match uniquely with neighbors' features.

DeLA is conceptually simple but powerful indeed.
We demonstrate its effectiveness on multiple point cloud tasks, including shape classification, 
scene semantic segmentation and object part segmentation.
In terms of both speed and accuracy, DeLA achieves the state-of-the-art (SOTA).

\section{Related work}

\paragraph{Point cloud networks.}

Existing works can generally be categorized into 2D-projection, voxelization or point based methods.
The first line of work projects a point cloud to images and processes them with well tested 2D convolution networks~\cite{simpleview, mvcnn}
or pre-trained models~\cite{p2p}.
Due to occlusions, multi views are often required to make good predictions.
Similarly, voxelization based methods project a point cloud onto 3D grids, 
which is then processed with 3D convolutions~\cite{octnet, mink}.
This approach preserves much finer 3D information, 
but local details would require voxelization of higher resolution to retain.
Point based methods operate directly on points and inter-point spatial relations.
Although the pioneering work PointNet~\cite{PointNet} processes the point cloud as a whole,
subsequent works following PointNet++~\cite{PointNet++} mostly focus on delicate local aggregation operators~\cite{KPConv, RSCNN, PointCNN, PointTransformer, PAConv} and hierarchical learning.
Our work falls under this category and discusses a way to capture inter-point relations economically through decoupling the local aggregation process.

\paragraph{Lightweight Local Aggregation.}
A typical local aggregation process consists of  neighboring feature gathering, processing and aggregation.
Authors in~\cite{gm2mg} demonstrate the order of gathering and processing can be switched to save computation 
from $O\left(edge\right)$  to $O\left(vertex\right)$ with equivalent expressivity,
when edge and center point features are processed by one linear layer, and aggregated with max-pooling.
PosPool~\cite{PosPool} shows using raw relative coordinates directly as weights to sum neighboring features can yield quite good performance under
a deep residual network setting.
APP-Net~\cite{APPNet} utilizes a two stage aggregation strategy where point features are pushed to and pulled from nearby auxiliary points.
This greatly reduces the cost of K nearest neighbor searching and memory operations.
Though its performance on large scenes is not tested.
\cite{effignn} is a work more related to ours, where the authors suggest to keep a powerful first layer that operates on raw point clouds, while simplifying the rest aggregation layers. 
We here point out the key is to encode necessary spatial information that distinguishes points at a local level.
Re-encoding is required for subsequent layers after downsampling, and the encoding layer can be cheap.

\paragraph{Decoupled learning.}
Decomposing a system into collaborative subsystems is likely to reduce internal complexity and strengthen configurability.
As a guiding ideology it is widely applicable.
Multifunctional operators can be decomposed into functionally similar operators:
a classical convolution can be decomposed into pointwise convolutions~\cite{NIN} and depthwise separable convolutions~\cite{Depthwise} or into spatially orthogonal colvolutions with lower dimensionalities~\cite{CylindricalAsy}.
Some works decouple loss functions to improve the quality of supervisory signals, such as decoupled weight decay~\cite{AdamW} and standalone distribution regularization~\cite{DistributionReg}.
Besides, penalizing over-collaboration among different parts of the model~\cite{Dropout, DropPath} can be seen as a type of soft decomposition.
We mainly focus on the simplification of local aggregation operators specific to point cloud learning, but the idea also guides model design and regularization setting.

\section{Method}

In this section, we first give a general description of local aggregation and briefly discuss some limitations of traditional approaches.
Then we detail the decoupling of local aggregation and elaborate our design of DeLA.

\subsection{Preliminary}

Local aggregation in point cloud learning functions as convolutions in image models to fuse information from nearby elements.
Unlike pixels though, points do not fall in a regular pattern. 
So local aggregation needs to explicitly model spatial relations  rather than simply encode them in parameters (like convolutions).
A typical local aggregation operation can be formalized as:
\begin{equation}
\label{eq:gla}
    \mathbf{y}_i = f\left(\left\{\left(\mathbf{x}_j, \mathbf{p}_j - \mathbf{p}_i \right) | j \in \mathcal{N}_i \right\}, \mathbf{x}_i\right),
\end{equation}
where neighbor features $\mathbf{x}_j$ and coordinates $\mathbf{p}_j$ relative to the center point $\mathbf{p}_i$ are first queried into a set,
which is then processed together with center feature $\mathbf{x}_i$ by some function $f$ to produce the output feature $\mathbf{y}_i$.
$\mathcal{N}_i$ denotes point $i$'s neighbors.
For simplicity and consistency, 
we use $j$ to denote a neighbor of $i$ throughout and omit that $j$ belongs to $\mathcal{N}_i$ hereafter.
Many classic aggregation operators take this form~\cite{PointNet++, PointCNN, RSCNN, KPConv, PosPool, PAConv, PointTransformer}.

Normally one such complete operation requires a computational complexity of at least $O\left(qkC\right)$ to map 
raw coordinates to some internal representation and mix with point features, 
where $k$ is neighbor count,  $C$ is channel dimension and $q$ is some method related variable likely in the same
order of magnitude as $k$ or $C$.
For example, in KPConv~\cite{KPConv}, coordinates are used to calculate weights for all kernel-neighbor pairs, which is cheap.
The weights are used to sum neighbor features for each kernel, which takes $O\left(qkC\right)$, where $q$ is the count of kernel points.
While in RS-Conv~\cite{RSCNN}, coordinates are mapped to $C$ with a shared MLP ($O\left(qkC\right)$, $q$ is hidden dimension)
and mixed through element-wise multiplication ($O\left(kC\right)$).
Additionally, to support advanced operations, intermediate feature maps like gathered neighbor features can't be optimized away,
which would consume extra memory that's $k$ times the size of input feature map. 
Both could limit the model from scaling up.

PosPool~\cite{PosPool} is a representative operator of lightweight incorporation, 
where raw coordinates or other easily calculable quantities are used directly as weights to average neighbor features:
\begin{equation}
\label{eq:pospool}
    y_{ik} = Avg\left\{ x_{jk} c_{ijk} \right\},
\end{equation}
where $k$ denotes the $k^{th}$ channel and $c_{ijk}$ is some quantity calculable from relative coordinates.
As a local aggregation operator, PosPool alone can only deliver limited functionality.
Specifically, it could suffer from incomplete information and structural bias.
Firstly, if we consider \autoref{eq:pospool} as a linear system with respect to $\mathbf{c}_{ij}$,
then it is only solvable when dimension is large enough and point $i$ has perfect information about all neighbor features $\mathbf{x}_j$, 
which is apparently not possible as these features are the ones to be gathered.
Secondly, as $\mathbf{c}_{ij}$ is not learnable, strong bias can be introduced. 
In the vanilla version, for example, where the 3 components of coordinates are used as weights, 
nearby neighbors tend to be ignored.

\subsection{Decoupled local aggregation}

We first introduce the concept of spatial encoding, 
with which we show a complete local aggregation operator can be decomposed into primitive ones.

\subsubsection{Spatial encoding}

\paragraph{Definition.}

We define spatial encoding $\mathbf{s}$ to be any point feature if relative coordinates can be decoded at a local level:
\begin{equation}
\label{eq:spatial_encoding}
    \left\{ \mathbf{p}_j - \mathbf{p}_i \right\} = decode\left( \left\{ \mathbf{s}_j \right\}, \mathbf{s}_i \right).
\end{equation}
This condition ensures that spatial information is locally intact centered on point $i$, 
which is demanded for clear local aggregation.
$decode$ here is an {\bf equivariant} set function: each $\mathbf{p}_j - \mathbf{p}_i$ corresponds to one $\mathbf{s}_j$.
The actual  decoding process may vary depending on information content.

\paragraph{Absolute spatial encoding.}

The most straightforward option is to use points' absolute  coordinates.
Such encodings can be cheaply formed.
However, absolute coordinates break translation invariance
and features learned in one coordinate range can't generalize well to another,
which is especially problematic under scenes of varying sizes.
Besides, absolute encodings may be too primitive to provide useful inductive biases, 
as no relations are contained.

\paragraph{Relative spatial encoding.}

An alternative is to encode relative coordinates:
\begin{equation}
\label{eq:rel_se}
    \mathbf{s}^{rel}_i = encode\left\{ \mathbf{p}_j - \mathbf{p}_i \right\}.
\end{equation}
Note the input here is different from the output of \autoref{eq:spatial_encoding} in that no order is required here.
The local point cloud centered on $i$ (including the neighbors of $i$ and their neighbors) can be seen as a directed graph and an edge from $m$ to $n$ exists only if $n$ is a neighbor of $m$.
An undirected graph can be derived if only bidirectional edges are kept.
Then $\mathbf{s}^{rel}$ is a valid spatial encoding if two conditions are met:
(1) Edges are all unique in the undirected graph, so any edge determines the relation between two points.
(2) Point $i$'s neighbors are connected in the undirected graph, so all neighbors can be located from $i$.
These conditions can be safely assumed to be true.
For condition (1), a specific edge has measure zero and is highly unlikely to collide with another in length and direction.
For condition (2), it's always possible to utilize a larger neighbor size for encoding than decoding to connect all neighbors. Though this is normally not necessary.

\subsubsection{Operator decomposition}

We consider the decomposition of $f$ in \autoref{eq:gla}.
Assume spatial encoding is already contained in point features, 
then by \autoref{eq:spatial_encoding}, 
relative coordinates can be decoded from point features at a local level:
\begin{equation}
    \left\{\left(\mathbf{x}_j, \mathbf{p}_j - \mathbf{p}_i \right)\right\} = decode\left(\left\{\mathbf{x}_j\right\}, \mathbf{x}_i\right).
\end{equation}
So there exists a function $f'$ that takes simplified inputs and outputs $\mathbf{y}_i$:
\begin{equation}
    \mathbf{y}_i = f'\left(\left\{\mathbf{x}_j\right\}, \mathbf{x}_i\right).
\end{equation}
Then, by the universal approximation theory on set functions~\cite{PointNet}, 
we embed the content of a set into a vector with transformation $g$ and max pooling $M$, and then extract with $h$:
\begin{equation}
\label{eq:decomdone}
    \mathbf{y}_i = h\left( M\left\{ g\left( \mathbf{x}_j\right)  \right\}, \mathbf{x}_i \right).
\end{equation}
This set embedding step may utilize other aggregation operators as well (average pooling~\cite{deepsets} for example).
Transformations $g$ and $h$ may be modelled by pointwise MLPs.
As all operations in \autoref{eq:decomdone} are either pointwise or primitive, the decomposition is complete.

To summarize, for simple operators to carry out fine functionalities, three conditions should be met:
(1) Encode essential spatial information in point features beforehand.
(2) Interleave feature transformations into pooling operations to both prepare the message sent and extract the message received.
(3) Use skip connections or concatenations to preserve the source feature after pooling.

\subsection{DeLA}

DeLA is a working instantiation of decoupled local aggregation.
Details on model architecture are presented below,
along with the regularization we employ to enhance local awareness,
and a brief analysis on model complexity.

\subsubsection{Architecture}
\label{sec:architecture}

\begin{figure*}[t]
  \centering
   \includegraphics[width=1.\textwidth]{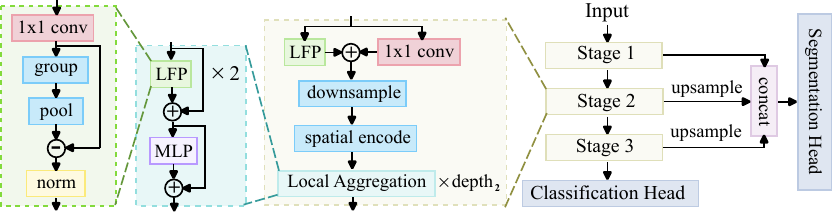}
   \caption{Architecture of DeLA.
   \textcircled{+} and \textcircled{-} denote element-wise addition and subtraction.
   }
   \label{fig:dela}
\end{figure*}

The general architecture of DeLA is given in \autoref{fig:dela}.
For the ease of understanding, we detail the design choices in a bottom-up style.

\paragraph{LFP.}

The local feature propagation (LFP) layer is illustrated in the leftmost of \autoref{fig:dela}.
We use edge max-pooling~\cite{DeepGCNs} for information communication.
The operation is identical to pooling followed by subtraction:
\begin{equation}
    max\left\{ \mathbf{x}_j \right\} - \mathbf{x}_i = max\left\{ \mathbf{x}_j - \mathbf{x}_i \right\}.
\end{equation}
Compared to pooling alone, this achieves a cheap second-order interaction, which is potentially beneficial for relation learning.
The preceding linear projection is used for feature recomposition, which improves the diversity of sent signals especially when skip connections are used.
The normalization helps stabilize the statistics after pooling.

\paragraph{Local Aggregation.}

We follow the recent trend in image model design~\cite{MetaFormer}, where a token mixer and a two-layer MLP with skip connections are taken as a unit and stacked to form the main block. 
Different in our context, we empirically find the need for feature transformation in point cloud tasks is not as strong, and thus two LFP layers are used with one MLP in a unit.
It is also inspired by the analysis in AIR~\cite{air}, that we don't necessarily need to use as many transformation layers as propagation.

\paragraph{Spatial encode.}

We use relative spatial encoding as in \autoref{eq:rel_se}.
The encoding process is modelled by a small PointNet~\cite{PointNet}, which is a three-layer MLP followed by max-pooling.
In the first stage, other inputs like color and normal are also gathered from neighbors and processed by this pointnet.
Points' height is also used as in PointNext~\cite{PointNext}, 
which may help DeLA form some global perception in early stages, as DeLA is otherwise purely translation invariant by design.
The dimension of this PointNet is set low, and a pointwise convolution is applied for spatial encoding to match dimension except in the first stage.
Then the encoding is normalized and serves as the initial feature embedding or added to main route.
A two-layer MLP with skip connection follows to prepare features for aggregation.

\paragraph{Downsample.}

Before point subsampling, a pointwise convolution is used to lift channels, 
combined with a LFP layer to reduce information loss, as shown in the middle of \autoref{fig:dela}.
Points are sampled with farthest point sampling (FPS)~\cite{PointNet++} on small scaled tasks 
or grid subsampling~\cite{msn, KPConv} on scene segmentation tasks.
Efficiency is our main concern, as FPS possesses a quadratic complexity while grid subsampling performs in linear time.
Moreover, as grid subsampling runs on CPU, computation can be done concurrent with model execution that utilizes GPU and little delay is incurred.
In our implementation of grid subsampling, we utilize an algorithm with two hash tables to optimize cache hit ratio.
This is based on our empirical observation that points adjacent in storage also tend to be spatially adjacent, 
thereby increasing the likelihood of them being in the same grid and processed in cache.

\paragraph{Task head.}

For classification, we lift the channels of the last stage's output and send the max-pooled features to a three-layer classifier, 
with dropout before the last linear layer.

For segmentation, the outputs of all stages are upsampled directly to the input resolution using nearest-neighbor interpolation.
These upsampled features are then concatenated and fed into a two-layer classifier.
We choose this strategy because the nearest neighbor is most likely located in the local structure centered on the target point,
and may thus carry the most clear semantic meaning.
To reduce overfitting, we randomly drop features from higher stages.
Equivalently, we replace upsample-concat-linear with linear-upsample-add in actual implementation, which saves computation.
In object part segmentation, where the shape category is given, we form a class embedding with it and concatenate it before the classifier.

\subsubsection{Regularization}

Even though spatial encoding is provided to the network,
it may not necessarily utilize this information as we intend.
To enhance DeLA's local awareness and reduce potential ambiguity during local aggregation,
we employ a regularization term that predicts the relative coordinates with neighbor points.
Specifically, at the end of each stage, 
we randomly select a neighbor for each point and compute the difference between their features.
This difference is fed into a two-layer head to get the prediction,
which is compared to the ground truth to calculate mean standard error loss.

For this term to have a relatively consistent regularization strength across different tasks and datasets, 
we normalize the ground truth to unit variance, and average losses from all stages.
Additionally, we decay this term exponentially to a negligible magnitude $\lambda$.
This ensures that the regularization yields strong guidance at the beginning, 
without adversely affecting the main learning course as training progresses.

\subsubsection{Complexity}

For a specific stage, denote neighbor count, channel dimension,
dimension of the PointNet in spatial encoding and stage depth as $k$, $C$, $C_p$ and $d$.
Then the overall complexity for local aggregation is that of spatial encoding $O\left( kC^2_p \right)$,
plus pointwise convolution $O\left( dC^2 \right)$, 
plus neighbor max-pooling $O\left( dkC \right)$.
$C_p$ is often much smaller than $C$ and we roughly have $kC^2_p \approx C^2$.
Therefore the amortized complexity for one local aggregation is basically $O\left( C^2 + kC \right) = O\left( C^2 \right)$.

\section{Experiments}

\subsection{Experimental setup}

We evaluate DeLA on five commonly adopted datasets:
S3DIS~\cite{S3DIS} and ScanNet v2~\cite{scannet} for scene semantic segmentation,
ShapeNetPart~\cite{shapenet} for object part segmentation,
and ScanObjectNN~\cite{ScanObjectNN} and ModelNet40~\cite{modelnet} for shape classification.
Network configurations on five datasets are listed in \autoref{tab:dela_config}.
For data augmentation, we use random scaling, feature (color or normal) dropping and color auto contrasting whenever applicable.
Random sampling is used for classification tasks during training.
Point resampling augmentation is applied on ScanObjectNN following ~\cite{PointNext}.
Other augmentation configurations are listed in \autoref{tab:dela_aug}.

All experiments are implemented with PyTorch~\cite{pytorch} and performed on one RTX 4090 with 24GB memory.
We train all models with cross entropy loss with label smoothing~\cite{labelsmooth}.
The optimizer is AdamW~\cite{AdamW} with cosine annealing and weight decay of 0.05.
Stochastic depth~\cite{DropPath} is applied to local aggregation blocks and the preceding MLP.
Normalization and activation are set to BatchNorm~\cite{batchnorm} and GELU~\cite{gelu}.
The final regularization strength $\lambda$ is set to 3e-3 for all tasks.
More detailed hyperparameter settings can be checked in our code.

For scene semantic segmentation, following the common practice,
we grid subsample the point cloud before sending to the network.
During training, a predefined maximum number of points are taken with a center crop and the rest are discarded.
During testing, the whole scene is processed.
The grid size and maximum number of points are set to 0.04m, 30000 and 0.02m, 80000 for S3DIS and ScanNet v2.
To form a complete prediction, we interpolate the predictions with nearest-neighbor interpolation and average several predictions.
The stochasticity comes solely from grid subsampling, and does not improve performance compared to making predictions for every individual point.
For ScanNet v2, estimated normal vectors are utilized as extra input, and test time augmentation is also employed following~\cite{pt2}.

We report the performance of the best run on these datasets, which is the common practice.
For ablation studies, we report mean $\pm$ standard of 3 random runs.

\begin{table}[t]
  \centering
  \caption{Network configurations on five datasets. FLOPs are counted per 1024 points.}
  \label{tab:dela_config}
  \begin{tabular}{lccccc}
    \toprule
    Dataset                                  & Dimension        & Depth      & K-neighbors       &      Params      &       FLOPs \\
    \midrule
    S3DIS~\cite{S3DIS}                       & 64-128-256-512       & 2-2-4-2  &   24              &     7.0M        &       0.96G  \\
    ScanNet v2~\cite{scannet}               &  64-96-160-288-512    & 2-2-2-4-2 &  24              &     8.0M        &      0.81G   \\
    ShapeNetPart~\cite{shapenet}            &  96-192-320-512       & 2-2-2-2  &    20            &      7.5M         &     1.85G   \\
    ScanObjectNN~\cite{ScanObjectNN}         & 96-192-384       & 2-2-2      &     24              &     5.3M        &       1.5G  \\
    ModelNet40~\cite{modelnet}              & 96-192-384       & 2-2-2      &     20              &     5.3M        &       1.44G  \\
    \bottomrule
  \end{tabular}
\end{table}
\begin{table}[t]
  \centering
  \caption{Data augmentations on five datasets.}
  \label{tab:dela_aug}
  \begin{tabular}{lcccccccc}
    \toprule
    Dataset                                   & Vertical rotation    &   Jittering       &  Height translation &  Elastic distortion \\
    \midrule
    S3DIS~\cite{S3DIS}                        &  \checkmark          &   \checkmark      &                     &                     \\
    ScanNet v2~\cite{scannet}                 &  \checkmark          &                   &                     &   \checkmark         \\
    ShapeNetPart~\cite{shapenet}              &                      &   \checkmark      &      \checkmark     &                    \\
    ScanObjectNN~\cite{ScanObjectNN}          &  \checkmark          &                   &                     &                    \\
    ModelNet40~\cite{modelnet}                  &                   &                   &    \checkmark        &                   \\
    \bottomrule
  \end{tabular}
  
\end{table}

\subsection{Experimental results}

\begin{table}[t]
	\centering
	\caption{{\small Semantic segmentation results on S3DIS Area 5.
    	Throughput (Thr.) (instances per second) is evaluated with an input of 16x15000 points.
            Class IoUs of ceiling, floor, wall and beam are omitted.
    	}}
        \label{tab:s3dis}
	\resizebox{1.0\textwidth}{!}{
		\begin{tabular}{ l  c c c c  c c c c c c c c c}
			\toprule 
			Method & OA & mAcc & mIoU & Thr.   & col. & wind. & door & table & chair & sofa & book. & board & clutter \\
			\midrule
			PTv1~\cite{PointTransformer} & 90.8 & 76.5 & 70.4 & 112  & 38.0 & 63.4 & 74.3 & 89.1 & 82.4 & 74.3 & 80.2 & 76.0 & 59.3 \\
                 FPT~\cite{FPT} & - & 77.6 & 71.0 & 96  & 53.8 & \textbf{71.2} & 77.3 & 81.3 & 89.4 & 60.1 & 72.8 & 80.4 & 58.9\\
			PointNeXt~\cite{PointNext} & 91.0 & 77.2 & 71.1 & 94  & 37.7 & 59.3 & 74.0 & 83.1 & 91.6 & 77.4 & 77.2 & 78.8 & 60.6\\
			PointMixer~\cite{pointmixer} & - & 77.4 & 71.4 & 69  & 43.8 & 62.1 & 78.5 & 90.6 & 82.2 & 73.9 & 79.8 & 78.5 & 59.4\\
			StratifiedFormer~\cite{StratifiedTransformer} & 91.5 & 78.1 & 72.0 & 71 & 46.1 & 60.0 & 76.8 & 92.6 & 84.5 & 77.8 & 75.2 & 78.1 & 64.0\\
			DU-Net~\cite{EdgeEnSu} & - & - & 72.2 & 121 & 40.0 & 60.7 & \textbf{82.7} & 90.8 & 83.1 & 78.5 & 83.5 & 75.9 & 64.1\\
                PTv2~\cite{pt2}        &  91.6  &  78.0   & 72.7  & 83  &  34.4    &   64.7  &  77.9 &  \textbf{93.1} & 84.4 & 77.3 & \textbf{86.3} & \textbf{84.5} & 62.2 \\
			\midrule
			DeLA (ours) & \textbf{92.2} & \textbf{80.0} & \textbf{74.1} & \textbf{468} & \textbf{55.1} & 61.4 & 76.0 & 84.4 & \textbf{93.2} & \textbf{85.7} & 79.4 & 80.8 & \textbf{65.2} \\
			\bottomrule
	\end{tabular}}
\end{table}

\begin{table}[t]
    \begin{minipage}{.4\textwidth}
    \centering
    \small
    \caption{Semantic segmentation results on ScanNet v2 validation set.}
    \label{tab:scannetv2}
    \begin{tabular}{l c c}
    \toprule
    Method  &   Params  &   mIoU    \\
    \midrule
    PTv1~\cite{PointTransformer}                    &   7.8M    &   70.6    \\
    FPT~\cite{FPT}                                  &   37.9M   &   72.1    \\
    MinkowskiNet~\cite{mink}                        &   37.9M   &   72.2    \\
    LargeKernel3D~\cite{largekernel3d}              &   40.2M   &   73.5    \\
    StratifiedFormer~\cite{StratifiedTransformer}   &   18.8M   &   74.3    \\
    PTv2~\cite{pt2}                                 &   11.3M   &   75.5    \\
    OctFormer~\cite{OctFormer}                      &   39M     &   75.7    \\
    \midrule
    DeLA (ours)                                     &   8.0M    &   75.9    \\
    \bottomrule
\end{tabular}
    \end{minipage}
    \hspace{.03\textwidth}
    \begin{minipage}{.5\textwidth}
    \centering
    \small
    \caption{Part segmentation results on ShapeNetPart. * denotes voting is used.}
    \label{tab:shape}
    \begin{tabular}{lcc}
    \toprule
    Method  & ins. mIoU & cat. mIoU\\
    \midrule
    Point Transformer~\cite{PointTransformer} & 86.6 & 83.7 \\
    PointMLP~\cite{pointmlp} & 86.1 & 84.6\\
    PVT~\cite{pvt} & 86.6 & -\\
    CurveNet*~\cite{CurveNet} & 86.8 & -\\
    PointNeXt-S*~\cite{PointNext} & 86.7 & 84.2\\
    PointNeXt-S (C=64)*~\cite{PointNext} & 86.9 & 85.2\\
    PointNeXt-S (C=160)*~\cite{PointNext} & 87.1 & 85.4\\
    DU-Net~\cite{EdgeEnSu} & 86.7 & 84.5\\
    DU-Net*~\cite{EdgeEnSu} & 87.1 & 85.2\\
    SPoTr*~\cite{spotr}     & 87.2  & 85.4 \\
    \midrule
    DeLA (ours) & 87.0 & 85.8\\
    DeLA (ours)* & 87.5  & 86.0\\
    \bottomrule
\end{tabular}

    \end{minipage}
\end{table}

\subsubsection{Semantic segmentation}

We perform semantic segmentation on two in-door datasets: S3DIS~\cite{S3DIS} and ScanNet v2~\cite{scannet}.
S3DIS consists of 271 rooms in six areas with 13 semantic categories.
ScanNet v2 is comparatively larger and contains 1513 scenes with 20 semantic categories for training and evaluation.
Following the common practice~\cite{pt2, StratifiedTransformer}, 
we use area 5 for testing on S3DIS and the official train/evaluation split on ScanNet v2.

Results of our method on these two datasets are listed in \autoref{tab:s3dis} and \autoref{tab:scannetv2} along with other SOTA methods.
On S3DIS, DeLA is 4\textasciitilde6x faster than other methods, 
while outperforming the previous SOTA Point Transformer v2~\cite{pt2} by 1.4\% in mIoU.
On ScanNet v2, DeLA also achieves the best validation performance with a relatively small number of learnable parameters.
The gain on S3DIS is larger than on ScanNet v2, 
we hypothesis the reason may be DeLA's simplicity, 
as S3DIS has smaller amount of data and is more prone to overfitting.
Though it's not clear to what extent DeLA could persist such advantage.
On the other hand, DeLA is shown to perform better on smaller objects like chair, sofa and clutter.
It could be that the actual local aggregation operator is simple, and the neighborhood is not large.
Therefore the effective receptive field may be limited.

\subsubsection{Part segmentation}

ShapeNetPart~\cite{shapenet} contains around 17000 3D samples with 50 part categories from 16 object categories.
We list results on this dataset in \autoref{tab:shape}.
Even without voting, DeLA already surpasses most of the methods.
In terms of category mIoU, DeLA has a larger improvement (+0.6\% compared to PointNeXt-S (C=160)~\cite{PointNext}) and
reaches 86\%.
This may imply that DeLA is less prone to overfitting.

\subsubsection{Shape classification}

\begin{table}[t]
    \begin{minipage}{.45\textwidth}
    \centering
    \small
    \caption{Classification results on ScanObjectNN (PB T50 RS).}
    \label{tab:scanobj}
    
\begin{tabular}{lccc}
    \toprule
    Method  & OA & mACC  & FLOPs\\
    \midrule
    PointNet~\cite{PointNet} & 75.2 & 71.4  & 1.0G\\
    PointNet++~\cite{PointNet++} & 86.2 & 84.4 & 1.7G\\
    DGCNN~\cite{dgcnn} & 86.1 & 84.3 & 4.8G\\
    PointMLP~\cite{pointmlp} & 87.7 & 86.4 & 31.4G\\
    PointNeXt~\cite{PointNext} & 88.2 & 86.8 & 1.6G\\
    SPoTr~\cite{spotr}         & 88.6 & 86.8 & - \\
    \midrule
    DeLA (ours) & 90.4  & 89.3 & 1.5G\\
    \bottomrule
\end{tabular}

    \end{minipage}
    \hspace{.03\textwidth}
    \begin{minipage}{.45\textwidth}
    \centering
    \small
    \caption{Classification results on ModelNet40.}
    \label{tab:modelnet}
    
\begin{tabular}{lccc}
    \toprule
    Method  & OA & mACC  & FLOPs\\
    \midrule
    PointNet++~\cite{PointNet++} & 93.0 & 90.7 & 1.7G\\
    CurveNet~\cite{CurveNet} & 93.8 & - & -\\
    PointMLP~\cite{pointmlp} & 94.1 & 91.3 & 31.4G\\
    PointNeXt~\cite{PointNext} & 94.0 & 91.1 & 6.5G\\
    PVT~\cite{pvt} & 94.1 & - & 1.9G\\
    \midrule
    DeLA (ours) & 94.0 & 92.2 & 1.44G\\
    \bottomrule
\end{tabular}
\label{tab:mdnt}

    \end{minipage}
\end{table}

We test our method on ScanObjectNN~\cite{ScanObjectNN} and ModelNet40~\cite{modelnet} for this task.
ScanObjectNN is a challenging real-world dataset generated from indoor scenes.
It consists of around 15000 samples with backgrounds and occlusions from 2902 unique objects  with 15 categories.
We use the hardest variant (PB\_T50\_RS) of this dataset.
ModelNet40 is a synthetic dataset comprised of around 12000 samples with 40 categories.
The points are uniformed sampled from CAD models, so there is much less noise compared to ScanObjectNN.
We report the results without voting in \autoref{tab:scanobj} and \autoref{tab:modelnet}.

As is shown, DeLA is again highly performant and efficient.
With only 1.5GFLOPs, DeLA exceeds 90\% overall accuracy on ScanObjectNN.
This is the first time a model can achieve this performance without voting or any extra training data.
DeLA also has the least difference between overall accuracy and mean class accuracy on both ScanObjectNN and ModelNet40, indicating good balance.

\subsection{Ablation studies}

\begin{table}[t]
    \begin{minipage}{.45\textwidth}
    \centering
    \small
    \caption{Effect of spatial encoding.}
    \label{tab:spatial_encoding}
    
\begin{tabular}{ccccc}
    \toprule
    Rel-first  & Rel-rest & Absolute & cat. mIoU\\
    \midrule
    \checkmark & \checkmark & height & 85.4$\pm$0.4\\
    \checkmark &            & height & 85.0$\pm$0.1\\
    \checkmark & \checkmark &        & 85.0$\pm$0.2\\
    \checkmark & \checkmark & \checkmark &  85.0$\pm$0.2  \\
    \checkmark &             & \checkmark &  85.3$\pm$0.3\\
               &            & \checkmark &  84.8$\pm$0.4\\
    \bottomrule
\end{tabular}
    \end{minipage}
    \hspace{.03\textwidth}
    \begin{minipage}{.45\textwidth}
    \centering
    \small
    \caption{Effect of aggregation operator.}
    \label{tab:agg_op}
    \begin{tabular}{lc}
    \toprule
    Operator  & mIoU\\
    \midrule
    no pre-conv & 72.4$\pm$0.3  \\
    max         & 72.1$\pm$0.1  \\
    edge max    & 73.5$\pm$0.5  \\
    average     & 67.8$\pm$0.1  \\
    edge average & 70.2$\pm$0.2 \\
    PosPool~\cite{PosPool}  & 72.8$\pm$0.3 \\
    edge PosPool & 71.1$\pm$0.3 \\
    \bottomrule
\end{tabular}
    \end{minipage}
    \vspace{0.4cm}
    \centering
    \small
    \caption{Effect of regularization strength.}
    \label{tab:reg_str}
    \begin{tabular}{lccccccc}
    \toprule
    Strength    &   0     &     1e-4        &       3e-4       &    1e-3    &   3e-3    &   1e-2    &       1   \\
    \midrule
    mIoU        & 72.8$\pm$0.3 & 73.1$\pm$0.4 & 73.4$\pm$0.4  & 73.3$\pm$0.3 & 73.5$\pm$0.5 & 73.2$\pm$0.2 & 72.5$\pm$0.3 \\
    \bottomrule
\end{tabular}
\end{table}

\subsubsection{Effect of spatial encoding}

To examine the effect of different spatial encodings,
we experiment with varying combinations on ShapeNetPart~\cite{shapenet}.
Because category mIoU is more discriminating here,
we report it in \autoref{tab:spatial_encoding}.
Rel-first denotes the relative spatial encoding layer in the first stage of DeLA,
and Rel-rest denotes  relative spatial encoding layers in subsequent stages.
Absolute means taking absolute coordinates as an extra input for the first spatial encoding layer,
and height means only up axis is taken.

Starting with the baseline, we can see that removing subsequent spatial encoding harms performance, which is expected.
Each time points are subsampled, their neighbors change, so new relations must be re-learned.
Reducing height also harms, indicating global information at early stages helps.
It is interesting that taking full absolute coordinates leads to inferior performance.
We hypothesis the reason to be overfitting, which is supported by the following configuration.
When Rel-rest is reduced, absolute coordinates becomes necessary in relations learning, which likely reduces overfitting.
Using absolute encoding alone produces the worst performance, likely due to lacking of useful relative priors.

To summarize, relative encoding works better.
While global information helps, height seems to suffice.

\subsubsection{Effect of actual aggregation operator.}

The Local Feature Propagation layer in \autoref{fig:dela} is ablated to examine the effect of operators that
carry out actual feature fusion.
We conduct this experiment on S3DIS~\cite{S3DIS} and report the results in \autoref{tab:agg_op}.
The first configuration ablates the preceding pointwise convolution.
The rest configurations replace the pooling operation.
An edge version means to perform pooling on edges (difference between point features) instead of point features.

The performance drop of the first configuration validates the importance of feature recomposition before information mixing.
As for pooling operations,
both max pooling and edge operations are shown beneficial.
PosPool~\cite{PosPool} demonstrates strong potential, but the performance deteriorates with edge operations.
It is likely because PosPool has already embedded strong relation priors.
Utilizing other cheap delicate operators for better relation learning might be worth investigating, though is not the focus of this paper.

\subsubsection{Effect of regularization strength.}

We list result on S3DIS~\cite{S3DIS} of different settings of the final regularization strength $\lambda$ in \autoref{tab:reg_str}.
The setting of $\lambda$ is generally insensitive, so long as it's not too large that impacts fitting or too small that has no effect.
Other decaying strategies are likely applicable.
Considering its insensitiveness, it may not worth too much resources to tune.

\section{Limitions and future work}

Firstly, decoupled local aggregation may not work well with shallow networks.
When the channel dimension is low, the cost of spatial encoding may need more aggregation operations to amortize.
Also, as basic operations are too primitive, they may need to be stacked deep enough to function well.
Secondly, as decoupled local aggregation targets local aggregation, its instantiations like DeLA lack long range interactions.

In future works, we plan on investigating:
(1) More economic strategies for spatial encoding.
(2) Advanced pooling operations that works well with decoupling for better relation learning.
(3) Expanding the receptive field.

\section{Conclusion}

In this work, we propose to decouple explicit relation learning from local aggregation,
where the key is to encode necessary  spatial information that distinguishs points at a local level.
We theoretically prove its validity and present an instantiation: DeLA, along with a regularization term to help enhance local awareness.
Extensive experiments on multiple point cloud datasets demonstrate the effectiveness of our proposed method.


\bibliographystyle{plainnat}
\bibliography{dela}

\end{document}